\documentclass[11pt,letterpaper]{article}
\usepackage[margin=1in]{geometry}
\usepackage[T1]{fontenc}
\usepackage{lmodern}
\usepackage{amsmath,amssymb,amsthm,mathtools}
\usepackage{microtype}
\usepackage[colorlinks=true,linkcolor=blue,citecolor=blue,urlcolor=blue]{hyperref}
\usepackage{cleveref}
\newtheorem{theorem}{Theorem}[section]
\newtheorem{lemma}[theorem]{Lemma}

\newtheorem{corollary}[theorem]{Corollary}

\theoremstyle{definition}
\newtheorem{definition}[theorem]{Definition}
\newtheorem{remark}[theorem]{Remark}
\newcommand{\E}{\mathbb E}
\newcommand{\Prb}{\mathbb P}
\newcommand{\Var}{\operatorname{Var}}

\newcommand{\one}{\mathbf 1}

\newcommand{\U}{\mathsf U}
\newcommand{\K}{\mathsf{K}}
\newcommand{\Tr}{\operatorname{Tr}}
\newcommand{\eps}{\varepsilon}
\newcommand{\poly}{\operatorname{poly}}

\newcommand{\R}{\mathbb R}
\newcommand{\bits}{\{0,1\}}
\newcommand{\abs}[1]{\left|#1\right|}
\newcommand{\norm}[1]{\left\|#1\right\|}
\newcommand{\ip}[2]{\left\langle #1,#2\right\rangle}
\newcommand{\wh}[1]{\widehat{#1}}

\DeclareMathOperator{\err}{err}
\title{Non-Evolvability of Monotone DNF\\
under the Uniform Distribution}
\author{Mohammad-Hadi Foroughmand-Araabi \and Sama Goliaei}
\date{}
\hypersetup{
  pdftitle={Non-Evolvability of Monotone DNF under the Uniform Distribution},
  pdfauthor={Mohammad-Hadi Foroughmand-Araabi and Sama Goliaei}
}
\begin{document}
\maketitle
\begin{abstract}
We prove that monotone disjunctive normal forms (DNFs) are not evolvable under
the uniform distribution in Valiant's model, resolving a question posed by
Valiant (2009) and highlighted by Feldman, Lee, and Servedio (2011). At
classification loss $\eps_0=1/100$, no adaptive randomized learner in the
correlational statistical-query (CSQ) or general statistical-query (SQ) model
can learn $n$-variable monotone DNFs with at most $s$ canonical terms using a
query count and inverse tolerance bounded by a single polynomial in $n$ and
$s$. More strongly, we construct a sequence of hard distributions with
$s=n^{o(1)}$ canonical terms for which every learner achieving loss at most
$\eps_0$ with probability at least $2/3$ must satisfy
$Q\tau^{-2}=n^{\omega(1)}$. This bound allows unlimited computation, improper
real-valued hypotheses, and independent unlabeled examples.
The proof constructs a prior by randomly permuting the coordinates of a
disjunction of rare monotone gates. We choose the gates so that conditioning
on rejection is nearly product on every set of a prescribed number of
coordinates, yielding constant prior variance and a covariance operator of
superpolynomially small norm.
\end{abstract}

\section{Introduction and main results}\label{sec:intro}

Valiant~\cite{Val09} introduced the computational model of \emph{evolvability}
within the PAC learning framework~\cite{Val84} to study how complex Boolean
mechanisms can arise through random variation and selection guided only by
empirical performance estimates. In Valiant's framework, a representation
$r:\bits^n\to[-1,1]$ undergoes local randomized mutations to produce a
polynomial-size neighborhood of candidate offspring, and selection chooses among
candidates using empirical estimates of their correlation
$\mathrm{Perf}_f(r)=\E_{X\sim\mathcal D}[f(X)r(X)]$ with the target function
$f:\bits^n\to\{-1,1\}$. Feldman~\cite{Fel08,Fel09} proved that evolvability of a
concept class over a distribution $\mathcal D$ is characterized by learnability
in the \emph{correlational statistical-query} (CSQ) model~\cite{BFJ+94,Kea98}
over $\mathcal D$.

For unrestricted (non-monotone) DNF formulas under the uniform distribution
$\U_n$, non-evolvability follows directly from the fact that parity functions of
size $\Theta(\log s)$ can be written as $s$-term DNFs and form an exponentially
large orthogonal family~\cite{BFJ+94}. Parity functions, however, are maximally
non-monotone. Valiant~\cite{Val09} showed that monotone conjunctions are
evolvable under the uniform distribution and asked whether richer monotone
classes, such as monotone DNF formulas, are also evolvable under $\U_n$
(see also~\cite{Fel08,Fel09,KVV10}).

Establishing unconditional CSQ lower bounds for monotone Boolean classes under
$\U_n$ requires different techniques because monotonicity ties a variable's
influence to its degree-$1$ Fourier coefficient~\cite{BT96,Ser04}. On the
algorithmic side, Verbeurgt~\cite{Ver90} and Bshouty and Tamon~\cite{BT96}
showed that $s$-term monotone DNFs are learnable under $\U_n$ in
quasipolynomial complexity $n^{O(\log(s/\eps))}$~\cite{LMN93,Jac97,KS04},
Feldman~\cite{Fel12} obtained the stronger upper bound
$\widetilde O\bigl(n\cdot(s\log(s/\eps))^{O(\log(s/\eps))}\bigr)$, and
Servedio~\cite{Ser04} and Kanade, Valiant, and Vaughan~\cite{KVV10} gave
polynomial-time algorithms in restricted parameter regimes. On the lower-bound
side, Dachman-Soled et al.~\cite{DLM+14} proved cryptographic hardness for
learning monotone functions, and Feldman, Lee, and Servedio~\cite{FLS11}
identified the learnability of monotone DNF under the uniform distribution as
the motivating open problem for their investigation of shallow monotone
formulas. By composing read-once Tribes formulas with monotone majority and
amplification gates, Feldman, Lee, and Servedio~\cite{FLS11} proved
unconditional CSQ and non-evolvability lower bounds for monotone formulas of
depth $3$ and depth $4$. Expanding such depth-$3$ or depth-$4$ compositions into
depth-$2$ disjunctive normal form, however, blows up the number of terms
exponentially.

In this work, we prove an unconditional superpolynomial lower bound for learning
depth-$2$ monotone DNF formulas under the uniform distribution in the CSQ and SQ
models, and deduce that monotone DNF is not evolvable under $\U_n$. Our result
does not rule out polynomial-time learning of monotone DNF from ordinary labeled
examples.

\subsection{Main statistical-query lower bounds}

Let $X=(X_1,\ldots,X_n)\sim\U_n$ consist of independent fair bits. A
\emph{monotone DNF} on $\bits^n$ with at most $s$ terms is a Boolean function of
the form
\begin{equation}\label{eq:dnf}
 F(x)=\bigvee_{j=1}^{s'}\ \bigwedge_{i\in T_j}x_i,
 \qquad s'\le s,
\end{equation}
where all literals are positive. Redundant or subsumed clauses may be removed, so
$s$ always bounds the \emph{canonical} number of minimal positive conjunctions
(minterms) of $F$, rather than the size of an arbitrary DNF representation.
We write $f=2F-1\in\{-1,1\}$ for the signed target label.

In the \emph{correlational statistical-query} (CSQ) model under $\U_n$, the
learner submits a bounded query function $g:\bits^n\to[-1,1]$ and a tolerance
$\tau_g>0$, and the oracle returns any real number
\[
 v_g\in\bigl[\E_{X\sim\U_n}[f(X)g(X)]-\tau_g,\;
             \E_{X\sim\U_n}[f(X)g(X)]+\tau_g\bigr].
\]
The oracle may choose $v_g$ adversarially and adaptively based on the entire
query history. The learner may be randomized and adaptive, may draw fresh
unlabeled samples from $\U_n$, and may output an arbitrary improper hypothesis
$r:\bits^n\to[-1,1]$. For a real-valued hypothesis $r:\bits^n\to[-1,1]$ and
signed label $f=2F-1\in\{-1,1\}$, we define the classification loss
\begin{equation}\label{eq:real-error}
 \err(r,f)=\frac{1-\E_{X\sim\U_n}[r(X)f(X)]}{2}.
\end{equation}
When $r=2A-1\in\{-1,1\}$ is Boolean, $\err(r,f)=\Prb_{X\sim\U_n}[A(X)\neq F(X)]$
is the standard $0$--$1$ classification error.

\begin{theorem}[Main CSQ lower bound for monotone DNF]\label{thm:main}
Fix $\eps_0=1/100$. There is no learner which, on every $n$, every $s$,
and every monotone DNF with at most $s$ canonical terms, achieves loss
$\err(r,f)\le\eps_0$ with probability at least $2/3$ against every legal CSQ
oracle and whose query count and inverse tolerance are bounded by a single
polynomial in $n$ and $s$. More precisely, there exist positive integers $n_j\to\infty$,
term budgets $s_j=n_j^{o(1)}$, and a sequence of hard distributions supported on
$n_j$-variable monotone DNFs with at most $s_j$ canonical terms for which every
randomized adaptive learner making at most $Q_j$ correlational queries of
tolerances at least $\tau_j$ and achieving loss $\err(r,f)\le\eps_0$ with
probability at least $2/3$ for every target $f$ in the support of the hard
distribution and against every legal CSQ oracle must satisfy
\begin{equation}\label{eq:strong-main}
             Q_j\tau_j^{-2}=n_j^{\omega(1)}.
\end{equation}
The conclusion holds even when the learner is required only to succeed with
probability at least $2/3$ on average over a target drawn from the hard
distribution, has unlimited computation apart from its query budget, outputs
improper Boolean or real-valued hypotheses $r:\bits^{n_j}\to[-1,1]$, and uses
independent unlabeled samples.
\end{theorem}

Because the input distribution $\U_n$ is fixed and known, \cref{thm:main} also
applies to general statistical queries~\cite{Kea98,BFJ+94,Fel17}.

\begin{corollary}\label[corollary]{cor:sq}
In the setting of \cref{thm:main}, suppose the learner may make arbitrary
statistical queries $\psi:\bits^n\times\{-1,1\}\to[-1,1]$ with tolerance
$\tau_\psi\ge\tau_j$, receiving adversarial answers in
$[\E_{X\sim\U_{n_j}}[\psi(X,f(X))]-\tau_\psi,\E_{X\sim\U_{n_j}}[\psi(X,f(X))]+\tau_\psi]$.
Then every randomized adaptive improper learner achieving loss at most
$\eps_0=1/100$ with probability at least $2/3$ for every target in the hard
family and against every legal SQ oracle must satisfy
$Q_j\tau_j^{-2}=n_j^{\omega(1)}$.
\end{corollary}
\begin{proof}
Since $f(x)\in\{-1,1\}$, any query $\psi:\bits^n\times\{-1,1\}\to[-1,1]$
decomposes pointwise as
\[
 \psi(x,f(x))=a_\psi(x)+f(x)b_\psi(x),\qquad
 a_\psi(x)=\frac{\psi(x,1)+\psi(x,-1)}{2},\quad
 b_\psi(x)=\frac{\psi(x,1)-\psi(x,-1)}{2}.
\]
Both $a_\psi$ and $b_\psi$ take values in $[-1,1]$ and depend only on $x$.
Because $\U_{n_j}$ is known (and the learner in \cref{thm:main} is
computationally unrestricted), the learner can compute the target-independent
mean $\mu_\psi=\E_{X\sim\U_{n_j}}[a_\psi(X)]$ exactly, submit the correlational
query $b_\psi$ at tolerance $\tau_\psi$, and add $\mu_\psi$ to any legal CSQ
answer $v_{b_\psi}$ to obtain a legal SQ answer $v_\psi=\mu_\psi+v_{b_\psi}$.
(For a computationally bounded learner, $\mu_\psi$ can be estimated to within
$\tau_\psi/2$ from $O(\tau_j^{-2}\log(Q_j/\delta))$ fresh unlabeled samples,
reducing the effective CSQ tolerance by at most a factor of $2$.) Thus
\cref{thm:main} implies the claim.
\end{proof}

\begin{remark}\label[remark]{rem:quantifiers}
Under $\U_n$, $s$-term DNF is learnable with complexity
$\widetilde O\bigl(n\cdot(s\log(s/\eps))^{O(\log(s/\eps))}\bigr)$~\cite{Fel12}
(and $n^{O(\log(s/\eps))}$~\cite{Ver90,BT96}).
On the hard sequence in \cref{thm:main}, $s_j=n_j^{o(1)}$, so every fixed
polynomial in $n_j$ and $s_j$ is polynomial in $n_j$.
The lower bound therefore excludes polynomial complexity on this sequence.
It is compatible with the cited upper bounds: subpolynomial growth of
$s_j$ does not imply that
$(s_j\log s_j)^{O(\log s_j)}$ is polynomial in $n_j$.
\end{remark}

The version of \cref{thm:main} with Boolean hypotheses and success required
for every target and legal oracle has been formalized in Lean~4, including
the quantitative conclusion~\eqref{eq:strong-main}. The formalization and
build instructions are included in the ancillary files.

\subsection{Non-evolvability of monotone DNF}\label{subsec:evolvability}

In Valiant's evolvability framework~\cite{Val09,Fel08}, the performance of a
representation $r:\bits^n\to[-1,1]$ relative to a target $f:\bits^n\to\{-1,1\}$
under $\U_n$ is $\mathrm{Perf}_f(r)=\E_{X\sim\U_n}[f(X)r(X)]=1-2\err(r,f)$. A
concept class $\mathcal C$ is evolvable under $\U_n$ if there is an evolutionary
algorithm using representations $\mathcal R\subseteq[-1,1]^{\bits^n}$,
neighborhood size $M\le\poly(n,s,1/\eps)$, generation budget
$T\le\poly(n,s,1/\eps)$, and performance tolerance
$\tau_{\mathrm{evol}}\ge 1/\poly(n,s,1/\eps)$ (estimated either from
$s_{\mathrm{evol}}\le\poly(n,s,1/\eps)$ labeled examples per candidate as
in~\cite{Val09}, or via an adversarial performance oracle returning values in
$[\mathrm{Perf}_f(r')-\tau_{\mathrm{evol}},\mathrm{Perf}_f(r')+\tau_{\mathrm{evol}}]$
as in~\cite{Fel08,Fel09}) that outputs $r_T$ with
$\mathrm{Perf}_f(r_T)\ge 1-2\eps$ with probability at least $2/3$ for every
$f\in\mathcal C$.

\begin{corollary}\label[corollary]{cor:evolvability}
The class of monotone DNF formulas is not evolvable under the uniform
distribution $\U_n$, either in Valiant's empirical-sample model~\cite{Val09} or
in the adversarial-tolerance model~\cite{Fel08}. In the
adversarial-tolerance model at error $\eps_0=1/100$ ($\mathrm{Perf}_f(r)\ge 49/50$)
on the hard sequence $(n_j,s_j)$ with $s_j=n_j^{o(1)}$ of \cref{thm:main}, every
evolutionary algorithm $\mathcal E$ (even with unrestricted computation and
arbitrary representations $\mathcal R\subseteq[-1,1]^{\bits^{n_j}}$) that
succeeds over $T_j$ generations with neighborhood size $M_j$ and tolerance
$\tau_j$ with probability at least $2/3$ for every target in the hard family
and against every legal performance oracle must satisfy
\begin{equation}\label{eq:evol-quant}
 T_j (M_j+1) \tau_j^{-2} = n_j^{\omega(1)}
\end{equation}
(and hence $T_j M_j \tau_j^{-2} = n_j^{\omega(1)}$ whenever $M_j\ge 1$).
\end{corollary}
\begin{proof}
For the qualitative non-evolvability statement in Valiant's empirical-sample
model, Theorem~4.1 of Feldman~\cite{Fel08} proves that any concept class that is
evolvable over a distribution $\mathcal D$ is learnable over $\mathcal D$ by an
adaptive CSQ algorithm using $\poly(n,s,1/\eps)$ queries of tolerance at least
$1/\poly(n,s,1/\eps)$. By \cref{thm:main}, monotone DNF is not CSQ-learnable
under $\U_n$ with resources bounded by $\poly(n,s)$ at $\eps_0=1/100$, and
therefore is not evolvable under $\U_n$.

For the quantitative bound \cref{eq:evol-quant} in the adversarial-tolerance
model, each generation $k\in\{1,\ldots,T_j\}$ queries the performance
$\mathrm{Perf}_f(r')=\E_{X\sim\U_{n_j}}[f(X)r'(X)]$ of at most $M_j+1$
functions $r'\in N(r_{k-1})\cup\{r_{k-1}\}\subseteq[-1,1]^{\bits^{n_j}}$ up to
additive error $\tau_j$, and after $T_j$ generations outputs a representation
$r_{T_j}:\bits^{n_j}\to[-1,1]$ whose loss is
$\err(r_{T_j},f)=(1-\mathrm{Perf}_f(r_{T_j}))/2$. This is directly an adaptive
randomized CSQ learner making at most $Q_j=T_j(M_j+1)$ correlational queries of
tolerance $\tau_j$ and outputting a real-valued hypothesis $r_{T_j}$. Applying
\cref{thm:main} directly to $r_{T_j}$ (without any rounding step) yields
$T_j(M_j+1)\tau_j^{-2}=n_j^{\omega(1)}$.
\end{proof}

\section{Proof overview}\label{sec:overview}

For $S\subseteq[m]$, write $\chi_S(x)=\prod_{i\in S}(2x_i-1)$ and
$\wh{H}(S)=\E_{\U_m}[H(X)\chi_S(X)]$; let $\U_{m,z}$ denote the product
distribution in which each bit has mean $1/2+z/w$. We first relate learning
to the covariance of a prior on Boolean targets $H_\theta$. With
$h=\E_\theta H_\theta$, let
$\K=\E_\theta[(H_\theta-h)\otimes(H_\theta-h)]$,
$V=\Tr\K$, and $\kappa=\norm{\K}_{\mathrm{op}}$.
A deterministic reference oracle answers each correlational query with its
prior average. For any fixed learner randomness, its reference queries and
output hypothesis are independent of the target. Chebyshev's inequality
bounds the probability of departure from this transcript by
$4Q\kappa/\tau^2$. The reference output has mean loss at least $V$ and loss
variance at most $\kappa$, so its probability of attaining loss at most
$\eps<V$ is at most $\kappa/(V-\eps)^2$. This gives the adaptive lower bound
in \cref{lem:reference}.

To construct the prior, take a rare monotone DNF gate $G$ on $m$ variables
with acceptance probability $p$, and form the disjunction of
$B=\lfloor1/p\rfloor$ copies on disjoint blocks. A uniformly random
permutation of the $Bm$ coordinates determines the blocks and their internal
coordinate order. The resulting functions remain depth-$2$ monotone DNFs,
with at most $B$ times as many canonical terms as $G$. The needed condition
on $G$ is product contact through order $t$: every marginal of
$\nu=\mathcal L(X\mid G(X)=0)$ on at most $t$ coordinates approximates a
common product distribution with relative error $B^{-\omega(1)}$.
For two block partitions, the overlap of their rejection indicators is
controlled by the number of blocks of the second partition on which $G$
fires under $\nu^{\otimes B}$. A tuple of $j$ blocks sees an almost-product
distribution unless it meets some original block in more than $t$
coordinates. Such exceptional tuples have probability $B^{-t+o(1)}$;
even for them, the density bound
$d\nu^{\otimes B}/d\U_{Bm}\le(1-p)^{-B}=O(1)$ bounds their joint firing
probability by $O(p^j)$. Averaging factorial moments before applying
Bonferroni's inequalities at depth $\lceil(\log B)^2\rceil$ gives
$\norm{\K}_{\mathrm{op}}\le B^{-t/2+o(1)}$ and
$\Tr\K\to e^{-1}-e^{-2}>1/5$; see \cref{lem:shuffle}.

The remaining task is to construct gates with this contact property and
few terms. We state the required theorem in \cref{sec:combining} and use it
there to deduce the main lower bound. Its construction begins with finite
positive mixtures of
$e^{2z}(1-e^{-\lambda e^{2cz}})$ whose Taylor coefficients match
$1+4z$ through degree $t$, with independently prescribed small perturbations
in degrees $2,\ldots,t$. Here $c=O(\log(t+2))$.
A Gamma-integral representation, cancellation of its tails on a compact
interval, and Carath\'eodory's theorem give these mixtures in
\cref{sec:jets}. Each component is then represented by a monotone DNF pool
with $O_t(2^{\lceil cw\rceil})$ clauses of width $\lceil cw\rceil$ on
$\Theta_t(w^2)$ variables, multiplied by a disjoint width-$w$ trigger.
Monotone binary threshold selectors with $L=2w^2$ bits provide weights
at resolution $2^{-2w^2}$ using at most $L+1$ clauses
(\cref{sec:templates}).

To match the individual Fourier coefficients, we add paired groups of
monotone threshold functions anchored on each set $S$ of at most $t$
coordinates. Their expected response is triangular in the relation
$S\subseteq T$, allowing the coefficients indexed by $T$ to be adjusted
together with their averages at each level. Randomly embedded triggers can
be chosen with every pair intersecting in fewer than $w/4$ coordinates;
the Boolean disjunction then differs from the sum of its active templates
by at most $O_t(N^2 2^{-7w/4})$ in $L^1$, where
$N=w^{A_t}$ is the replication budget and $P=N2^{-w}$ is the firing scale.
Concentration and this overlap bound preserve the signs of the Fourier
residuals on opposite faces of the control cube. The
Poincar\'e--Miranda theorem, applied to their multilinear interpolation
on the selector grid, yields a grid vertex whose Boolean gate has
product-contact error at most $2^{-w^2}$. Expanding its constituents gives
at most $p_w^{-O(\log(t+2))}$ canonical terms. The details are in
\cref{sec:calibration}.

\section{From small covariance to SQ hardness}\label{sec:covariance}

For a prior distribution $\Pi$ on Boolean targets $H_\theta:\bits^n\to\bits$ and
signed labels $f_\theta=2H_\theta-1$, define
\begin{align*}
 h(x)&=\E_\theta H_\theta(x),\\
 \K&=\E_\theta[(H_\theta-h)\otimes(H_\theta-h)]
       \quad\text{on }L^2(\U_n),\\
 V&=\Tr\K=\E_X\bigl[h(X)(1-h(X))\bigr],\qquad
 \kappa=\norm{\K}_{\mathrm{op}},
\end{align*}
where $(u\otimes u)g=\ip{u}{g}u$ with respect to the inner product on
$L^2(\U_n)$.

\begin{lemma}\label[lemma]{lem:reference}
Suppose $0<\eps<V$. For any adaptive randomized learner making at most $Q$
bounded signed-label CSQs, each of tolerance at least $\tau$ on all legal
executions, there is a legal oracle policy under which its prior-average
probability of outputting a Boolean or real-valued hypothesis
$r:\bits^n\to[-1,1]$ with loss $\err(r,f_\theta)\le\eps$ is at most
\begin{equation}\label{eq:reference}
       \frac{4Q\kappa}{\tau^2}+\frac{\kappa}{(V-\eps)^2}.
\end{equation}
The bound holds for arbitrary bounded, computationally unrestricted test
functions and arbitrary improper real-valued output hypotheses.
\end{lemma}
\begin{proof}
Fix the learner's internal random seed (including any independent unlabeled
samples) and consider the deterministic \emph{reference transcript} obtained by
answering each correlational query $g:\bits^n\to[-1,1]$ with its prior mean
$\bar v_g=\E_\theta\E_X[f_\theta(X)g(X)]=\E_X[(2h(X)-1)g(X)]$.

Truncate the reference simulation before any $(Q+1)$-st query or any query
of tolerance below $\tau$, assigning it an arbitrary fixed output if it
is truncated; write $g_1,\ldots,g_{Q'}$ for the resulting queries, where
$Q'\le Q$ and $\tau_{g_i}\ge\tau$.
If truncation occurs, the preceding reference answers are incompatible
with every target by the resource bounds on legal executions, so for
every target a legal oracle must already have departed from the
reference path.

At a fixed random seed, the reference queries $g_1,\ldots,g_{Q'}$ are
deterministic and independent of $\theta$. For each query $g_i$ with
$\norm{g_i}_2\le 1$, the target-dependent expectation
$\E_X[f_\theta(X)g_i(X)]-\bar v_{g_i}=2\ip{H_\theta-h}{g_i}$ has mean zero and
variance
\[
 \Var_\theta\bigl(\E_X[f_\theta(X)g_i(X)]\bigr)
 =4\ip{g_i}{\K g_i}\le 4\kappa\norm{g_i}_2^2\le 4\kappa.
\]
By Chebyshev's inequality and a union bound over $1\le i\le Q'$, a prior-random
target $\theta\sim\Pi$ satisfies
$\abs{\E_X[f_\theta(X)g_i(X)]-\bar v_{g_i}}>\tau\le\tau_{g_i}$ for at least one
$i\le Q'$ with probability at most $4Q\kappa/\tau^2$.

Define the oracle policy for a true target $H_\theta$ to return the reference
answer $\bar v_g$ whenever $\abs{\E_X[f_\theta(X)g(X)]-\bar v_g}\le\tau_g$, and
to return the exact expectation $\E_X[f_\theta(X)g(X)]$ otherwise. This policy
is always legal, and unless the departure event above occurs, the learner's
entire execution follows the target-independent reference path and produces a
target-independent hypothesis $r:\bits^n\to[-1,1]$.

Write $A(x)=(1+r(x))/2\in[0,1]$, so $r(x)=2A(x)-1$. Since
$f_\theta(x)=2H_\theta(x)-1\in\{-1,1\}$, we have the pointwise identity
\[
 \frac{1-r(x)f_\theta(x)}{2}
 =A(x)(1-H_\theta(x))+(1-A(x))H_\theta(x).
\]
Taking expectations over $X\sim\U_n$ and $\theta\sim\Pi$, the prior mean loss of
the target-independent hypothesis $r$ satisfies
\begin{align*}
 \E_\theta\err(r,f_\theta)
 &=\E_X\bigl[A(X)(1-h(X))+(1-A(X))h(X)\bigr]\\
 &\ge\E_X\min\bigl(h(X),1-h(X)\bigr)
 \ge\E_X\bigl[h(X)(1-h(X))\bigr]=V.
\end{align*}
The loss $\err(r,f_\theta)=\E_X[A(X)]-\ip{r}{H_\theta}$ is an affine functional
of $H_\theta$ with linear coefficient $-r$ satisfying $\norm{r}_2\le\norm{r}_\infty\le 1$,
so its variance over $\theta\sim\Pi$ is
\[
 \Var_\theta\bigl(\err(r,f_\theta)\bigr)
 =\ip{r}{\K r}\le\kappa\norm{r}_2^2\le\kappa.
\]
By Chebyshev's inequality,
\[
 \Prb_\theta\bigl(\err(r,f_\theta)\le\eps\bigr)
 \le\Prb_\theta\bigl(\abs{\err(r,f_\theta)-\E_\theta\err(r,f_\theta)}\ge V-\eps\bigr)
 \le\frac{\kappa}{(V-\eps)^2}.
\]
Combining the reference-departure event with the small-loss event on the
reference path and averaging over the learner's internal randomness gives
\cref{eq:reference}.
\end{proof}

\section{Constructing hard families by random permutations}\label{sec:shuffle}

Consider a monotone Boolean gate $G:\bits^m\to\bits$ with acceptance probability
$p=\Prb_{\U_m}(G=1)$ and rejection probability $q=1-p$. Denote the distribution
of $X\sim\U_m$ conditioned on rejection by $\nu=\mathcal L(X\mid G(X)=0)$.

\begin{definition}\label[definition]{def:contact}
The gate $G$ has \emph{$(t,\xi,\alpha)$-product contact} if, for every subset
$S\subseteq[m]$ with $|S|\le t$ and every assignment $a\in\bits^S$,
\begin{equation}\label{eq:contact}
 \nu(X_S=a)=\big(1+\delta_{S,a}\big)
          \alpha^{|a|}(1-\alpha)^{|S|-|a|},
 \qquad |\delta_{S,a}|\le\xi.
\end{equation}
\end{definition}

\begin{lemma}\label[lemma]{lem:shuffle}
Let $t\ge2$ be fixed. Suppose $G:\bits^m\to\bits$ is monotone with $p\to0$,
$B=\lfloor1/p\rfloor$, $m=B^{o(1)}$, $0<\alpha\le1/2$, $m|\alpha-1/2|=o(1)$, and
$G$ has $(t,\xi_B,\alpha)$-product contact with $\xi_B=B^{-\omega(1)}$. Form a
prior $\Pi$ on $Bm$-variable monotone DNFs by choosing a uniformly random
bijection $\pi:[B]\times[m]\to[Bm]$ (equivalently, partitioning the $Bm$
coordinates into $B$ ordered blocks $\pi(1),\ldots,\pi(B)$ of size $m$ whose
internal coordinates are also uniformly randomly ordered) and setting
\[
 F_\pi(X)=\bigvee_{b=1}^B G(X_{\pi(b)}),\qquad
 N_\pi=1-F_\pi,
\]
where $X_{\pi(b)}=(X_{\pi(b,1)},\ldots,X_{\pi(b,m)})$. Let $\K$ be the
covariance operator of $N_\pi$ (and hence of $F_\pi$) under $\Pi$. Then
\begin{equation}\label{eq:shuffle-result}
  \Tr\K\longrightarrow e^{-1}-e^{-2}>0,
  \qquad
  \norm{\K}_{\mathrm{op}}\le B^{-t/2+o(1)}.
\end{equation}
If $G$ has $s_G$ positive conjunction terms, every $F_\pi$ is a monotone DNF
with at most $B s_G$ canonical terms.
\end{lemma}

\begin{proof}
Set $p_\alpha=\Prb_{\mathrm{Bern}(\alpha)^m}(G=1)$. The product distributions
$\U_m$ and $\mathrm{Bern}(\alpha)^m$ have pointwise density ratio bounded
between $(1-2|\alpha-1/2|)^m$ and $(1+2|\alpha-1/2|)^m$; since
$m|\alpha-1/2|=o(1)$, this implies $p_\alpha/p=1+o(1)$. Because $G$ is monotone
and $\alpha\le1/2$, coupling $\mathrm{Bern}(\alpha)^m$ below $\U_m$ also gives
$p_\alpha\le p$.

Fix a reference permutation $\pi$ and draw an independent uniformly random
permutation $\pi'$. Conditioned on the event $N_\pi(X)=1$, the $B$ blocks of
$\pi$ are independent and each block tuple $X_{\pi(b)}$ is distributed according
to $\nu$. For every $x\in\bits^{Bm}$, the conditional probability mass
under $\nu^{\otimes B}$ satisfies $\Prb(X=x\mid N_\pi(X)=1)\le q^{-B}2^{-Bm}$,
so the density of the conditional law relative to $\U_{Bm}$ is bounded
everywhere by $q^{-B}=O(1)$. Write the overlap kernel as
\[
 J(\pi,\pi')=\ip{N_\pi}{N_{\pi'}}
 =q^B\Prb\bigl(N_{\pi'}(X)=1\bigm| N_\pi(X)=1\bigr).
\]
For a subset $I\subseteq[B]$ of $j$ block indices of $\pi'$, call $I$
\emph{exceptional} if the union $U_I=\bigcup_{b'\in I}\pi'(b')$ of its $jm$
coordinates intersects some block $\pi(b)$ in at least $t+1$ coordinates, and
\emph{regular} otherwise. Since any fixed $(t+1)$-element subset of $\pi(b)$ is
contained in the uniformly random $jm$-subset $U_I$ with probability
$\binom{jm}{t+1}/\binom{Bm}{t+1}\le (j/B)^{t+1}$, a union bound over the $B$
blocks of $\pi$ and their $\binom{m}{t+1}$ subsets of size $t+1$ gives
\begin{equation}\label{eq:bad-tuples}
 \Prb_{\pi'}(I\text{ is exceptional})
 \le B\binom{m}{t+1}\left(\frac{j}{B}\right)^{t+1}
 = B^{-t+o(1)},\qquad j\le L_B:=\lceil(\log B)^2\rceil.
\end{equation}
When $I$ is regular, $U_I$ meets each block $\pi(b)$ in a subset $S_b$ of size
$|S_b|\le t$. Because the blocks of $\pi$ are independent under $N_\pi=1$ and at
most $jm$ blocks have $S_b\neq\varnothing$, \cref{eq:contact} implies that the
joint conditional distribution of the $jm$ coordinates in $U_I$ differs
multiplicatively from $\mathrm{Bern}(\alpha)^{jm}$ by at most
$(1+\xi_B)^{jm}-1=B^{-\omega(1)}$. Thus, for every regular $I$,
\[
 \Prb\bigl\{\forall b'\in I,\; G(X_{\pi'(b')})=1\bigm| N_\pi(X)=1\bigr\}
 = p_\alpha^j\bigl(1+B^{-\omega(1)}\bigr).
\]
For every $I$ (including exceptional ones), the $j$ blocks of $\pi'$ in
$I$ are disjoint and therefore independent under the unconditional uniform
distribution $\U_{Bm}$, where each fires with probability $p$. Since conditioning
on $N_\pi(X)=1$ increases the probability of any event by at most the density
bound $q^{-B}$, we have for every fixed $\pi,\pi'$ and every $I$:
\begin{equation}\label{eq:bad-domination}
 \Prb\bigl\{\forall b'\in I,\; G(X_{\pi'(b')})=1\bigm| N_\pi(X)=1\bigr\}
 \le q^{-B}p^j=O(p^j).
\end{equation}

Let $Y=\sum_{b'=1}^B G(X_{\pi'(b')})$ denote the number of blocks of $\pi'$ that
fire. For a fixed $\pi'$, the $j$-th factorial moment
$\E[\binom{Y}{j}\mid N_\pi=1,\pi']$ is the sum over $|I|=j$ of the firing
probabilities above. Separating the sum into regular and exceptional subsets $I$
and averaging the absolute discrepancy over the random choice of $\pi'$,
\cref{eq:bad-tuples,eq:bad-domination} yield
\begin{equation}\label{eq:factorial-error}
 \E_{\pi'}\abs{\E\left[\binom{Y}{j}\Bigm| N_\pi=1,\pi'\right]
                -\binom{B}{j}p_\alpha^j}
 \le \binom{B}{j}p^j\bigl(B^{-\omega(1)}+O(1)\Prb_{\pi'}(I\text{ is exceptional})\bigr)
 \le \frac{B^{-t+o(1)}}{j!}
\end{equation}
uniformly for $0\le j\le L_B$, since $Bp=1+o(1)$.

By Bonferroni's inequalities, for any integer $L_B\ge 1$, the indicator
$\one_{\{Y=0\}}$ is sandwiched between the alternating partial sums of
$\binom{Y}{j}$ truncated at $L_B-1$ and $L_B$, so the truncation error is bounded
by $\binom{Y}{L_B}$. Under $N_\pi=1$, \cref{eq:bad-domination} bounds its
expectation by $q^{-B}\binom{B}{L_B}p^{L_B}=O(1/L_B!)=B^{-\omega(1)}$, and the
same bound holds for the binomial series of $(1-p_\alpha)^B$. Summing
\cref{eq:factorial-error} over $0\le j<L_B$ therefore gives
\begin{equation}\label{eq:overlap-concentration}
  \E_{\pi'}\abs{J(\pi,\pi')-q^B(1-p_\alpha)^B}
   \le B^{-t+o(1)}.
\end{equation}
For $h=\E_\pi N_\pi$, we have
\[
 \Tr\K=\E_\pi\norm{N_\pi}_2^2-\norm{h}_2^2
       =q^B-\E_{\pi,\pi'}J(\pi,\pi')
       \longrightarrow e^{-1}-e^{-2}>0,
\]
since $q^B\to e^{-1}$ and $q^B(1-p_\alpha)^B\to e^{-2}$. Finally, the squared
Hilbert--Schmidt norm $\norm{\K}_{\mathrm{HS}}^2$ equals the mean square of the
doubly centered kernel
$J(\pi,\pi')-\E_\rho J(\pi,\rho)-\E_\rho J(\rho,\pi')+\E J$. Since double
centering is an orthogonal projection in $L^2(\Pi\times\Pi)$ and
$J(\pi,\pi'),q^B(1-p_\alpha)^B\in[0,1]$,
\begin{align*}
 \norm{\K}_{\mathrm{op}}^2
 \le\norm{\K}_{\mathrm{HS}}^2
 \le\Var_{\pi,\pi'}J(\pi,\pi')
 \le\E_{\pi,\pi'}\bigl(J(\pi,\pi')-q^B(1-p_\alpha)^B\bigr)^2
 \le B^{-t+o(1)}.
\end{align*}
Taking square roots gives $\norm{\K}_{\mathrm{op}}\le B^{-t/2+o(1)}$.
\end{proof}

\section{The rare-gate theorem and the main lower bound}\label{sec:combining}

We state the technical theorem and deduce \cref{thm:main}.
The theorem is proved in \cref{sec:jets,sec:templates,sec:calibration}.

\begin{theorem}[Rare monotone gate theorem]\label{thm:gate}
For every fixed integer $t\ge2$ there exist constants $C_t,A_t>0$ and, for
all sufficiently large integers $w$, a monotone DNF
$G_w:\bits^{m_w}\to\bits$ such that, with $p_w=\E_{\U_{m_w}}[G_w]$,
\begin{align}\label{eq:gate-properties}
 m_w&\le C_t w^2,\qquad
 p_w=\Theta_t(w^{A_t}2^{-w}),\\
 s(G_w)&\le p_w^{-C\log(t+2)}\quad\text{for one absolute constant }C>0,\notag
\end{align}
and the conditional distribution $\mathcal L(X\mid G_w(X)=0)$ has
$(t,\xi_w,\alpha_w)$-product contact with
\begin{equation}\label{eq:gate-precision}
  \xi_w\le 2^{-w^2},\qquad
  \alpha_w=\frac12+O_t(p_w w/m_w),\qquad
  \alpha_w\le\frac12.
\end{equation}
In particular, for $B=\lfloor1/p_w\rfloor$, we have $m_w=B^{o(1)}$,
$m_w|\alpha_w-1/2|=o(1)$, and $\xi_w=B^{-\omega(1)}$.
\end{theorem}

\begin{proof}[Proof of \cref{thm:main} assuming \cref{thm:gate}]
Fix an arbitrary polynomial resource exponent $D\ge 1$. Choose a fixed integer
contact order $t\ge 2$ large enough that
\begin{equation}\label{eq:degree-choice}
 \frac t2>10D\bigl(C\log(t+2)+10\bigr),
\end{equation}
where $C$ is the absolute constant in \cref{eq:gate-properties}. For all
sufficiently large $w$, let $G_w:\bits^{m_w}\to\bits$ be the monotone DNF gate
given by \cref{thm:gate}, and set $B=\lfloor 1/p_w\rfloor$. Since
$m_w\le C_t w^2$ and $p_w=\Theta_t(w^{A_t}2^{-w})$, we have
$B=2^{w-O_t(\log w)}$ and $m_w=B^{o(1)}$. By \cref{eq:gate-precision}, all
hypotheses of \cref{lem:shuffle} are satisfied. The resulting permutation prior
$\Pi$ on $n_0=Bm_w=B^{1+o(1)}$ variables has
\begin{equation}\label{eq:main-K}
 V=\Tr\K=e^{-1}-e^{-2}+o(1)>\frac15,
 \qquad
 \kappa=\norm{\K}_{\mathrm{op}}\le B^{-t/2+o(1)},
\end{equation}
and is supported on monotone DNFs with at most
\[
 s_0\le B\cdot s(G_w)\le B^{C\log(t+2)+1}
\]
canonical terms.

At loss threshold $\eps_0=1/100<1/5$, \cref{lem:reference,eq:main-K} imply that
any randomized adaptive CSQ learner achieving loss $\err(r,f)\le\eps_0$ with
probability at least $2/3$ using at most $Q$ queries of tolerance at least $\tau$
must satisfy
\[
 \frac{4Q\kappa}{\tau^2}+\frac{\kappa}{(V-\eps_0)^2}\ge\frac23
 \quad\Longrightarrow\quad
 \frac{Q}{\tau^2}\ge B^{t/2-o(1)}.
\]
If a learner had $Q,\tau^{-1}\le(n_0+s_0)^D$, then $Q/\tau^2\le(n_0+s_0)^{3D}\le B^{3D(C\log(t+2)+2)}$,
contradicting \cref{eq:degree-choice} for all sufficiently large $w$.

To obtain a single sequence $(n_j,s_j)$ with $s_j=n_j^{o(1)}$ satisfying
\cref{eq:strong-main}, set $h_t=C\log(t+2)+2$ and $r_t=\sqrt{t h_t}$, and pad
each $n_0$-variable target with unused coordinates to ambient dimension
$n=\lceil B^{r_t}\rceil$. Because the padding bits are independent fair coins and
the target does not depend on them, conditional expectation onto the active
$n_0$ coordinates does not increase the $L^2$ or $L^\infty$ norm of any test
function or real-valued hypothesis and preserves its inner product with every
target $H_\theta$, while the covariance operator on $L^2(\U_n)$ retains its
nonzero spectrum through the isometric embedding
$L^2(\U_{n_0})\hookrightarrow L^2(\U_n)$ (and therefore has the same trace $V$
and operator norm $\kappa$). In terms of the padded dimension $n$, we have
\[
 s_0\le B^{h_t}=n^{\sqrt{h_t/t}+o(1)},\qquad
 \frac{Q}{\tau^2}\ge B^{t/2-o(1)}=n^{\frac12\sqrt{t/h_t}-o(1)}.
\]
Choosing $t_j=j+2$ and selecting $w_j$ large enough that $n_j\ge j$ and the
fixed-$t_j$ asymptotic bounds hold at $w_j$ ensures $n_j\to\infty$,
$s_j=n_j^{o(1)}$, and $Q_j\tau_j^{-2}=n_j^{\omega(1)}$, completing the proof of
\cref{thm:main}.
\end{proof}

\section{Constructing positive mixtures with prescribed moments}\label{sec:jets}

We now begin the proof of \cref{thm:gate}. The first step is an analytic lemma
constructing finite positive mixtures of functions of the form
$e^{2z}(1-e^{-\lambda_j e^{2cz}})$ whose Taylor expansion at $z=0$ matches the
linear polynomial $1+4z$ through degree $t$, with prescribed perturbations in
degrees $2$ through $t$.

\begin{lemma}\label[lemma]{lem:radial}
For every fixed integer $t\ge2$, let $c=C_0\log(t+2)$, where $C_0$ is
a sufficiently large absolute constant; $C_0=10^6$ suffices. There exists
$\eta_t>0$ such that for each sign vector $\sigma\in\{-1,1\}^{t-1}$ there are finitely many
rates $\lambda_j\in(0,\infty)$ and nonnegative weights $a_{\sigma,0}$ and
$a_{\sigma,j}$, with total weight at most $2$, such that
\begin{equation}\label{eq:jet-profile}
 e^{2z}\left(a_{\sigma,0}
  +\sum_j a_{\sigma,j}\bigl(1-e^{-\lambda_j e^{2cz}}\bigr)\right)
 =1+4z+\eta_t\sum_{\ell=2}^t\sigma_\ell z^\ell
          +O_t(z^{t+1})
\end{equation}
near $z=0$. The weights and rates depend only on $t$ and $\sigma$, and not on
$w$.
\end{lemma}
\begin{proof}
Set $\beta=1/c$ and $y=e^{2cz}$. The unperturbed analytic function in $y$ is
\[
 H_*(y)=y^{-\beta}(1+2\beta\log y),\qquad
 e^{2z}H_*(e^{2cz})=1+4z,
\]
with derivative $H_*'(y)=\beta y^{-1-\beta}(1-2\beta\log y)$. By differentiating the Gamma
integral $\int_0^\infty\lambda^\beta e^{-y\lambda}\,d\lambda=\Gamma(1+\beta)y^{-1-\beta}$
with respect to $\beta$, we have
\[
 \int_0^\infty\lambda^\beta(\log\lambda-\psi(1+\beta))e^{-y\lambda}\,d\lambda
 =-\Gamma(1+\beta)y^{-1-\beta}\log y,
\]
where $\psi=\Gamma'/\Gamma$. The derivatives of $H_*'$ at $y=1$
through order $t-1$ coincide with the moments $\int_0^\infty(-\lambda)^j L(\lambda)W(\lambda)\,d\lambda$
of the signed density
\begin{equation}\label{eq:signed-gamma}
 L(\lambda)W(\lambda)\,d\lambda,
 \qquad W(\lambda)=\lambda^\beta e^{-\lambda},\quad
 L(\lambda)=\frac{\beta}{\Gamma(1+\beta)}
 \left[1+2\beta\big(\log\lambda-\psi(1+\beta)\big)\right].
\end{equation}

We replace this signed density on $(0,\infty)$ by a positive density supported
on the compact interval
\[
 a=(t+2)^{-32},\qquad b=100(t+2)^2
\]
with identical moments through degree $t-1$. On $[a,b]$, the elementary bounds
$|\psi(1+\beta)|\le2$ and $0.9\le\Gamma(1+\beta)\le1.1$ imply $\beta/2\le L(\lambda)\le2\beta$.
We claim that for every polynomial $r$ of degree at most $t-1$,
\begin{equation}\label{eq:tail}
 \left|\int_{(0,a)\cup(b,\infty)}r(\lambda)L(\lambda)W(\lambda)
       \,d\lambda\right|
 \le\frac{\beta}{8}\int_a^b |r(\lambda)|W(\lambda)\,d\lambda.
\end{equation}
To prove this, let $M_r=\max_{\lambda\in[0,1]}|r(\lambda)|$. Markov's polynomial
inequality on $[0,1]$ gives $\max_{[0,1]}|r'|\le2(t-1)^2M_r$, so there is a
subinterval of $[1/(16(t+1)^2),1]$ of length at least $1/(16(t+1)^2)$ on which
$|r(\lambda)|\ge M_r/2$. Since $W(\lambda)\ge e^{-1}/(16(t+1)^2)$ on that
subinterval,
\[
 \int_a^b|r(\lambda)|W(\lambda)\,d\lambda\ge\frac{M_r}{1500(t+1)^4}.
\]
On $(0,a)$, using $|r(\lambda)|\le M_r$ and $\beta\log(1/a)<10^{-3}$ gives an upper
bound of $6\beta aM_r$. On $(b,\infty)$, Chebyshev's growth estimate
$|r(\lambda)|\le(4\lambda)^tM_r$ for $\lambda\ge1$ together with $e^{-\lambda}$
gives an upper bound of $20\beta e^{-b/2}M_r$. With $a=(t+2)^{-32}$ and
$b=100(t+2)^2$, the sum of the two tail integrals is at most $(\beta/8)\int_a^b|r|W$,
proving \cref{eq:tail}.

Viewing the tail functional $r\mapsto\int_{(0,a)\cup(b,\infty)}rLW\,d\lambda$ on
the finite-dimensional subspace of degree-$(t-1)$ polynomials in
$L^1([a,b],W\,d\lambda)$, the Hahn--Banach theorem and the duality
$(L^1)^*=L^\infty$ yield a measurable function $h:[a,b]\to\R$ with
$\norm{h}_\infty\le \beta/8$ such that
\[
 \int_a^b r(\lambda)h(\lambda)W(\lambda)\,d\lambda
 =\int_{(0,a)\cup(b,\infty)}r(\lambda)L(\lambda)W(\lambda)\,d\lambda
 \qquad(\deg r\le t-1).
\]
Then $d\nu=(L+h)W\one_{[a,b]}\,d\lambda$ is a positive measure on $[a,b]$ with
 density $(L+h)\ge 3\beta/8>0$ whose moments through degree $t-1$ match those of
$L(\lambda)W(\lambda)\,d\lambda$ on $(0,\infty)$. Setting
\begin{equation}\label{eq:mu}
 d\mu(\lambda)=e^{\lambda}\lambda^{-1}d\nu(\lambda)
      =(L(\lambda)+h(\lambda))\lambda^{\beta-1}\one_{[a,b]}(\lambda)\,d\lambda,
\end{equation}
we have $\mu([a,b])\le 3(b^\beta-a^\beta)<1/8$ because $\beta\log(b/a)<10^{-3}$. Defining
\[
 a_0=1-\int_a^b(1-e^{-\lambda})\,d\mu(\lambda)>0,
\]
the function $H(y)=a_0+\int_a^b(1-e^{-\lambda y})\,d\mu(\lambda)$ satisfies
$H(1)=H_*(1)=1$ and $H^{(j)}(1)=H_*^{(j)}(1)$ for all $1\le j\le t$.

To obtain the signed perturbations $\eta_t\sum_{\ell=2}^t\sigma_\ell z^\ell$ in
\cref{eq:jet-profile}, observe that the Taylor coefficients of $e^{2z}H(e^{2cz})$
of orders $2\le\ell\le t$ at $z=0$ are a nonsingular triangular linear image of
the moments $\int_a^b\lambda^{j-1}e^{-\lambda}\,d\mu(\lambda)=\int_a^b\lambda^{j-1}\,d\nu(\lambda)$
for $1\le j\le t$ (with the $j=1$ moment fixed so that the linear coefficient in
$z$ remains $4$). Since the moment map $P\mapsto(\int_a^b\lambda^j P(\lambda)W(\lambda)\,d\lambda)_{j=0}^{t-1}$
has a strictly positive-definite Gram matrix on polynomials of degree at most
$t-1$, choosing $\eta_t>0$ sufficiently small allows every sign pattern
$\sigma\in\{-1,1\}^{t-1}$ to be realized by adding a polynomial perturbation to
$h$ of sup norm at most $\beta/8$. The resulting density $L+h$ remains at least
$\beta/4>0$ on $[a,b]$ with total $\mu$-mass below $1/4$, and $a_{\sigma,0}>0$ is
chosen so that the value at $z=0$ equals $1$.

Finally, by Carath\'eodory's theorem on the compact interval $[a,b]$, the
continuous measure $\mu_\sigma$ can be replaced by a finite positive atomic
measure supported on at most $t+3$ points $\lambda_j\in[a,b]$ having the exact
same values of $\int d\mu_\sigma$, $\int(1-e^{-\lambda})\,d\mu_\sigma$, and
$\int\lambda^j e^{-\lambda}\,d\mu_\sigma$ for $1\le j\le t$. This preserves the
Taylor expansion through order $t$ and completes the proof.
\end{proof}

\section{Realizing the mixtures as monotone DNFs}\label{sec:templates}

We next construct the monotone DNF building blocks that realize the positive
mixtures of \cref{lem:radial} as Boolean formulas. A \emph{trigger} of
width $w$ is the conjunction of $w$ distinct variables.

\begin{lemma}\label[lemma]{lem:pool}
Fix $t\ge 2$, $c>0$, $\lambda>0$, and an accuracy $\delta>0$. There exists a
constant $C=C(t,c,\lambda,\delta)>0$ such that, for every sufficiently large $w$,
on $M=\lceil Cw^2\rceil$ variables there is a monotone DNF $P_{w,\lambda}$ with
at most $\lfloor\lambda 2^{\lceil cw\rceil}\rfloor$ conjunctions of width
$\lceil cw\rceil$ satisfying, simultaneously for all $0\le j\le t$,
\begin{equation}\label{eq:pool-jet}
 \left|[z^j]\Prb_{\U_{M,z}}(P_{w,\lambda}=1)
       -[z^j](1-e^{-\lambda e^{2cz}})\right|<\delta.
\end{equation}
\end{lemma}
\begin{proof}
Let $v=\lceil cw\rceil$ and $R=\lfloor\lambda 2^v\rfloor$, and sample $R$
independent uniformly random $v$-subsets of $[M]$ as monotone conjunctions.
Conditioned on an input $X\in\bits^M$ having Hamming weight $K=\sum_{i=1}^M X_i$,
the probability over the random choice of the $R$ clauses that at least one
clause is satisfied by $X$ is
\[
 h_w(K)=1-\left(1-\frac{(K)_v}{(M)_v}\right)^R,
\]
where $(a)_v=a(a-1)\cdots(a-v+1)$. Under $X\sim\U_M$, write
$Z_w=(2K-M)/\sqrt M$. By the central limit theorem, $Z_w\Rightarrow Z\sim N(0,1)$
as $w\to\infty$ for fixed $C$, and uniformly on bounded values of $Z_w$,
\[
 2^v\frac{(K)_v}{(M)_v}
  \longrightarrow \exp\left(\frac c{\sqrt C}Z-\frac{c^2}{2C}\right).
\]
Hence $h_w(K)$ converges in distribution and in expectation to
\[
 h_C(Z)=1-\exp\left[-\lambda
              \exp\left(\frac{cZ}{\sqrt C}-\frac{c^2}{2C}\right)\right].
\]

To extract the Taylor coefficients in $z$ under the biased product measure
$\U_{M,z}$, we use the generating identity
\begin{equation}\label{eq:score}
 \sum_{j\ge0} z^j\left(\frac2w\right)^j
       \sum_{|T|=j}\chi_T(X)
       =\prod_{i=1}^M\left(1+\frac{2z}{w}\chi_i(X)\right)
       =\left(1+\frac{2z}{w}\right)^K\left(1-\frac{2z}{w}\right)^{M-K}.
\end{equation}
For each fixed $0\le j\le t$, the $z^j$ coefficient of the right-hand side
converges to $[z^j]\exp(2\sqrt C zZ-2Cz^2)$, a degree-$j$ Hermite polynomial in
$Z$ with uniformly integrable moments of all fixed orders. By Gaussian
exponential tilting,
\[
 \lim_{w\to\infty}[z^j]\E_{\U_{M,z}}\bigl[\E_{\mathrm{clauses}} P_{w,\lambda}\bigr]
 =[z^j]\E_Z\bigl[h_C(Z+2\sqrt C z)\bigr]
 =[z^j]\E_Z\left[1-e^{-\lambda
             \exp(2cz+cZ/\sqrt C-c^2/(2C))}\right].
\]
As $C\to\infty$, dominated differentiation under the expectation shows that these
derivatives converge to $[z^j](1-e^{-\lambda e^{2cz}})$ for $0\le j\le t$. We
fix $C$ large enough that all $t+1$ derivative errors are strictly below
$\delta/2$.

Let $A_j=[z^j]\E_{\U_{M,z}}P_{w,\lambda}$ and expose the $R$ independent
clauses successively. Replacing one clause changes $A_j$ by at most
$b_j=2^{1-v}(2/w)^j\binom{M}{j}$, so the corresponding martingale
differences $D_{r,j}$ satisfy $|D_{r,j}|\le b_j$ and
\[
 \Var(A_j)=\sum_{r=1}^{R}\E D_{r,j}^2
 \le Rb_j^2
 =O_{t,C,\lambda}(2^{-v}\poly(w))=o(1).
\]
Chebyshev's inequality and a union bound over $0\le j\le t$ give a
single deterministic pool satisfying \cref{eq:pool-jet} for all
sufficiently large $w$.
\end{proof}

\begin{lemma}\label[lemma]{lem:selector}
For every integer $L\ge 1$ and integer cutpoint $0\le k\le2^L$, the indicator
$S_k:\bits^L\to\bits$ of the set of $L$-bit strings whose binary numerical value
is at least $k$ is a monotone DNF with at most $L+1$ conjunctions. Uniformly
over $0\le k\le 2^L$, under the biased product law $\U_{L,z}$,
\begin{equation}\label{eq:selector-jet}
 \Prb_{\U_L}(S_k=1)=1-k2^{-L},\qquad
 \big|[z^j]\E_{\U_{L,z}}S_k\big|
       \le4(2/w)^j\quad(1\le j\le t).
\end{equation}
\end{lemma}
\begin{proof}
For $k=2^L$, no $L$-bit string has value at least $2^L$, so $S_{2^L}\equiv 0$ is
the empty disjunction ($0\le L+1$ terms) and satisfies \cref{eq:selector-jet}
trivially. For $0\le k<2^L$, an $L$-bit string $y\in\bits^L$ has binary value
strictly greater than the $L$-bit binary expansion of $k$ if and only if at the
most significant bit position $r$ where $y$ differs from $k$, the expansion of
$k$ has a $0$ and $y$ has a $1$ (and $y$ has $1$s at all more significant
positions where $k$ has a $1$). Taking the disjunction of these conjunctions
together with the conjunction of all $1$-bits of $k$ itself yields a monotone
DNF with at most $L+1$ terms. The accepted interval $[k,2^L-1]$
partitions into disjoint dyadic prefix subcubes, with at most two subcubes of
each codimension $r\ge 1$. A subcube of codimension $r$ has $\U_L$-measure
$2^{-r}$ and $z^j$ coefficient bounded in absolute value by
$2^{-r}\binom{r}{j}(2/w)^j$. Summing over $r\ge j$ using
$\sum_{r\ge j}2^{-r}\binom{r}{j}=2$ gives \cref{eq:selector-jet}.
\end{proof}

\begin{lemma}\label[lemma]{lem:generic}
Fix $t\ge 2$ and let $c,\eta_t$ be as in \cref{lem:radial}. On a block of
$m=M+w+L+t$ variables, where $M=\lceil C_t' w^2\rceil$ and $L=2w^2$, there is a
finite collection (depending only on $t$) of monotone DNF template shapes, each
consisting of a width-$w$ trigger conjunction multiplied by either $1$ or a
disjoint sparse pool from \cref{lem:pool}, whose normalized expected bias
expansions under $\U_{m,z}$ can be positively combined to approximate
\[
        1+4z+\eta_t\sum_{\ell=2}^t u_\ell z^\ell
\]
through degree $t$, uniformly over $u_{\mathrm{rad}}=(u_2,\ldots,u_t)\in[-1,1]^{t-1}$,
within any prescribed accuracy $\delta>0$ independent of $w$. The combination
weights are nonnegative, bounded by $1$, and Lipschitz in $u_{\mathrm{rad}}$
with Lipschitz constant depending only on $t$.
\end{lemma}
\begin{proof}
For each sign vertex $\sigma\in\{-1,1\}^{t-1}$, take the finite positive mixture
from \cref{lem:radial}. A width-$w$ trigger conjunction has normalized biased
acceptance
\[
  2^w\Prb_{\U_{w,z}}(\text{trigger}=1)
     =(1+2z/w)^w\longrightarrow e^{2z}
\]
together with all fixed derivatives at $z=0$. Multiplying the trigger by a
variable-disjoint sparse pool $P_{w,\lambda_j}$ from \cref{lem:pool} realizes
$e^{2z}(1-e^{-\lambda_j e^{2cz}})$ to any accuracy $\delta>0$ for $C_t'$
sufficiently large. For a general control vector
$u_{\mathrm{rad}}=(u_2,\ldots,u_t)\in[-1,1]^{t-1}$, the multilinear barycentric
weights
\[
   \theta_{\sigma,j}(u_{\mathrm{rad}})=a_{\sigma,j}
        \prod_{\ell=2}^t\frac{1+\sigma_\ell u_\ell}{2},
   \qquad
   \theta_{\sigma,0}(u_{\mathrm{rad}})=a_{\sigma,0}
        \prod_{\ell=2}^t\frac{1+\sigma_\ell u_\ell}{2}
\]
are nonnegative, bounded by $1$, and $O_t(1)$-Lipschitz, and their weighted sum
reproduces $1+4z+\eta_t\sum_{\ell=2}^t u_\ell z^\ell$ up to error $O_t(\delta)+o_w(1)$.
When a fractional weight is realized through a disjoint $L$-bit selector from
\cref{lem:selector} with $L=2w^2$, its degree-$0$ acceptance matches the weight
within $2^{-2w^2}$ and its higher derivatives are $O_t(w^{-1})=o_w(1)$ by
\cref{eq:selector-jet}.
\end{proof}

\section{Matching all low-order moments}\label{sec:calibration}

We now complete the proof of \cref{thm:gate} by combining the generic derivative
templates of \cref{lem:generic} with groups of monotone threshold functions that
control individual low-order Fourier coefficients, and solving the resulting
system of moment equations on a fine discrete selector grid.

\begin{proof}[Proof of \cref{thm:gate}]
Fix $t\ge 2$, and let $c,\eta_t$ be given by \cref{lem:radial}. Let
$M=\lceil C_t'w^2\rceil$, $L=2w^2$, and $m=M+L+w+t=\Theta_t(w^2)$.
Choose an integer $A_t$ sufficiently large in terms of $t$;
$A_t=100(t+2)$ suffices. Set $N=w^{A_t}$ and $P=N2^{-w}$.

\paragraph{Generic template lists.}
For each template shape in \cref{lem:generic}, sample a list of $N$ independent
uniformly randomly embedded copies on the $m$ variables (keeping the trigger,
pool, and $L$-bit selector variables disjoint within each copy). At average
Fourier control vector $u_{\mathrm{rad}}=(u_2,\ldots,u_t)\in[-1,1]^{t-1}$,
activate the first $\lfloor N\theta(u_{\mathrm{rad}})\rfloor$ copies in the list
completely, and activate the next copy through its $L$-bit monotone selector with
acceptance threshold rounded to an integer multiple of $2^{-L}=2^{-2w^2}$. In the
sum of expectations of the active copies, normalized by $P$, the degree-$t$
Taylor polynomial in $z$ under $\U_{m,z}$ is
\begin{equation}\label{eq:ideal-radial}
 1+4z+\eta_t\sum_{\ell=2}^t u_\ell z^\ell
       +O_t(\delta)+o_w(1).
\end{equation}

\paragraph{Groups of threshold functions.}
To control individual Fourier coefficients on subsets $S\subseteq[m]$ of size
$1\le\ell\le t$, define on $\bits^S$ the pair of functions
\[
 h_S^\pm(x_S)=\frac12+\frac1{8t}\sum_{i\in S}\chi_i(x)
                       \pm\frac1{16t}\chi_S(x).
\]
Flipping any coordinate $j\in S$ from $0$ to $1$ increases
$\frac{1}{8t}\sum_{i\in S}\chi_i(x)$ by $\frac{1}{4t}$ and changes
$\pm\frac{1}{16t}\chi_S(x)$ by at most $\frac{1}{8t}$, so $h_S^\pm(x_S)$
increases by at least $\frac{1}{8t}>0$. Since $|S|=\ell\le t$, we have
$|h_S^\pm(x_S)-\frac12|\le\frac{\ell}{8t}+\frac{1}{16t}\le\frac{3}{16}<\frac12$,
and $Dh_S^\pm(x_S)=8t+2\sum_{i\in S}\chi_i(x)\pm\chi_S(x)\in\{1,\ldots,D-1\}$ is
an integer for $D=16t$. Thus each threshold indicator
$H_{S,r}^\pm(x_S)=\one_{\{Dh_S^\pm(x_S)\ge r\}}$ ($1\le r\le D$) is a monotone
Boolean function on $\bits^S$, representable as a monotone DNF of at most
$2^\ell\le 2^t$ terms, and satisfies the exact decomposition
\begin{equation}\label{eq:superlevels}
 \sum_{r=1}^{D}H_{S,r}^\pm(x_S)=Dh_S^\pm(x_S).
\end{equation}
A \emph{threshold group} of sign $\pm$ on $S$ consists of $D$ monotone
constituents obtained by multiplying each $H_{S,r}^\pm(x_S)$ ($1\le r\le D$) by
an independently sampled width-$w$ trigger conjunction and an
independently embedded $L$-bit monotone selector on $[m]\setminus S$ (disjoint
from that constituent's trigger). All $D$ constituents in a threshold group
therefore have independent random trigger embeddings; they share only the
numerical cutpoint $k\in\{0,\ldots,2^L\}$ of their respective selectors.

For small positive constants $\rho_t,\rho_{t-1},\ldots,\rho_1>0$ chosen below
and each subset $S\subseteq[m]$ of size $\ell\in\{1,\ldots,t\}$, sample $K_\ell$
independent plus groups and $K_\ell$ independent minus groups, where
\begin{equation}\label{eq:Kell}
 K_\ell=\left\lfloor
     \frac{N\rho_\ell w^{\ell-t}}{\binom{m}{\ell}}
                 \right\rfloor,
 \qquad
 \gamma_\ell=D\frac1{16t}K_\ell2^{-w}=K_\ell2^{-w}.
\end{equation}
At control parameter $u_S\in[-1,1]$, activate an effective fraction
$(1+u_S)/2$ of the plus groups and $(1-u_S)/2$ of the minus groups for $S$
(using complete prefixes and at most one selector-weighted group per sign).

\paragraph{Exact expected Fourier response.}
Let $T\subseteq[m]$ with $|T|=r\le t$, and consider a completely active
constituent on $S$ with $|S|=\ell$ and $R=T\cap S$. Because the width-$w$ trigger
is chosen uniformly at random from $[m]\setminus S$, it contains $T\setminus S$
with probability $(w)_{r-|R|}/(m-\ell)_{r-|R|}$, so
\begin{equation}\label{eq:anchor-expectation}
 \E_{\mathrm{embedding}}\wh{\bigl[\text{trigger}\cdot H_{S,j}^\pm\bigr]}(T)
 =2^{-w}\wh{H}_{S,j}^\pm(R)
   \frac{(w)_{r-|R|}}{(m-\ell)_{r-|R|}}.
\end{equation}
When we sum over $1\le j\le D$ using \cref{eq:superlevels} and combine the plus
and minus lists with weights $(1+u_S)/2$ and $(1-u_S)/2$:
\begin{itemize}
\item For $|S|=\ell\ge 2$, the constant term $1/2$ and the singleton terms
  $\frac{1}{8t}\sum_{i\in S}\chi_i$ have identical coefficients in $h_S^+$ and
  $h_S^-$, so their dependence on $u_S$ cancels completely, leaving only the
  top character $\pm\frac{1}{16t}\chi_S$ when $R=S$ (i.e., $S\subseteq T$).
\item For $|S|=\ell=1$ ($S=\{i\}$), the constant term $1/2$ cancels between
  $h_{\{i\}}^+$ and $h_{\{i\}}^-$, while the perturbation
  $\pm\frac{1}{16t}\chi_{\{i\}}=\pm\frac{1}{16t}\chi_i$ is itself the singleton
  character on $S$ and produces the varying response when $\{i\}\subseteq T$.
\end{itemize}
In all cases $1\le\ell\le t$, the $u_S$-dependent expected Fourier response at
$T$ is given by the upper-triangular formula
\begin{equation}\label{eq:triangular}
 \one_{\{S\subseteq T\}}\gamma_\ell u_S
              \frac{(w)_{r-\ell}}{(m-\ell)_{r-\ell}},
\end{equation}
while the $u_S$-independent baseline sum over all $\binom{m}{\ell}$ subsets $S$
is symmetric under coordinate permutations.

\paragraph{Concentration and disjunction-versus-sum bound.}
Each embedded constituent has Fourier coefficients bounded in magnitude by
$2^{-w}$. Across all $\poly_t(w)$ list prefixes and subsets $|T|\le t$,
Hoeffding's inequality and a union bound ensure the existence of fixed embeddings
for which the sum $S_u(x)$ of the active template indicators and its
expectation $\bar S_u(x)$ over the random embeddings satisfy, uniformly over
the control cube,
\begin{equation}\label{eq:uniform-linear}
 \max_{|T|\le t}\left|\wh{S_u}(T)
                    -\wh{\bar S_u}(T)\right|
 \le C_t2^{-w}m^t\sqrt{N\log w}+C_t2^{-w}m^t.
\end{equation}
Because all $O_t(N)$ constituents across all generic and
threshold lists (including constituents within the same threshold group) have
independently sampled width-$w$ triggers, any two distinct triggers intersect in
at least $w/4$ coordinates with probability at most
$\binom{w}{\lceil w/4\rceil}(w/(m-t))^{\lceil w/4\rceil}=w^{-\Omega(w)}$. By a
union bound over all $O_t(N^2)$ constituent pairs, we may choose the embeddings
so that the triggers of every pair of distinct constituents intersect in fewer
than $w/4$ variables. Let $G(u)$ be the Boolean disjunction of all
active constituents at control vector $u$. Pointwise,
$0\le S_u-G(u)\le\binom{S_u}{2}$, and every pair of
distinct constituents fires simultaneously with probability at most
$2^{-(2w-w/4)}=2^{-7w/4}$. Thus, uniformly over $u$,
\begin{equation}\label{eq:OR-SUM}
   \E_{\U_m}|G(u)-S_u|
          \le C_tN^2 2^{-7w/4}.
\end{equation}
Dividing \cref{eq:uniform-linear,eq:OR-SUM} by $P=N2^{-w}$ or by
$\min_{\ell\le t}\gamma_\ell=\Omega_t(P w^{-2t})$ gives $o_w(1)$, since $A_t$ is sufficiently large and $N=w^{O_t(1)}$.

\paragraph{Solving the moment equations.}
For the Boolean disjunction $G(u)$, write its Fourier coefficients as
\[
 b_T=\wh{G(u)}(T),\quad p=b_\varnothing,\quad q=1-p,\quad
 \bar b_\ell=\binom{m}{\ell}^{-1}\sum_{|T|=\ell}b_T,
 \quad a=-\bar b_1/q.
\]
Uniformly over $u$, we have
\begin{equation}\label{eq:a-estimate}
  p=\Theta_t(P),\qquad
  |a|\le C_t Pw/m.
\end{equation}
For $2\le\ell\le t$, define the average Fourier residual at level $\ell$:
\begin{equation}\label{eq:radial-coordinate}
 R_\ell(u)=\frac{(2/w)^\ell}{P}
       \sum_{|T|=\ell}(b_T+q a^\ell).
\end{equation}
For $1\le\ell\le t$ and each subset $T\subseteq[m]$ with $|T|=\ell$, define the
individual Fourier residual:
\begin{equation}\label{eq:angular-coordinate}
 A_T(u)=\frac{b_T-\bar b_\ell}{\gamma_\ell}
          +\bar u_\ell,
 \qquad \bar u_\ell=\binom{m}{\ell}^{-1}
                         \sum_{|S|=\ell}u_S.
\end{equation}
The number of equations matches the dimension
$d=(t-1)+\sum_{\ell=1}^t\binom{m}{\ell}$ of the control cube
$[-1,1]^{d}$.

We now bound the perturbation of $R_\ell(u)$ and $A_T(u)$ around their diagonal
control coordinates $\eta_t u_\ell$ and $u_T$. First consider
how much a threshold constituent on $S$ ($|S|=r\le t$) can contribute to the
scaled degree-$\ell$ sum $(2/w)^\ell\sum_{|T|=\ell}\wh{H}(T)$. A completely active
threshold constituent $H=\text{trigger}\cdot H_{S,j}^\pm$ depends on at most
$w+r\le w+t$ coordinates and takes values in $\bits$ with
$\E_{\U_m}[H]\le 2^{-w}$. Thus, $\wh{H}(T)=0$ unless $T$ is contained in
its $(w+r)$-variable support, and $|\wh{H}(T)|\le 2^{-w}$ for each of the at most
$\binom{w+t}{\ell}$ such subsets $T$, so its total contribution to
$(2/w)^\ell\sum_{|T|=\ell}\wh{H}(T)$ is bounded in magnitude by
\begin{equation}\label{eq:support-radial-bound}
 \left(\frac2w\right)^\ell\binom{w+t}{\ell}2^{-w}
 =O_t(2^{-w}).
\end{equation}
Similarly, a selector-weighted threshold constituent is the product of a disjoint
$L$-bit selector with $H$, so by \cref{eq:selector-jet,eq:support-radial-bound}
its degree-$\ell$ contribution is also $O_t(2^{-w})$. Since there are at most
$2D\sum_{r=1}^t\binom{m}{r}K_r\le O_t\bigl(N\sum_{r=1}^t\rho_r w^{r-t}\bigr)
\le O_t\bigl(N\sum_{r=1}^t\rho_r\bigr)$ threshold constituents in total across
all subsets $S$, dividing by $P=N2^{-w}$ shows that all threshold groups combined
perturb $R_\ell(u)$ by at most $C_t'\sum_{r=1}^t\rho_r$. By
\cref{eq:a-estimate}, the product-bias correction in \cref{eq:radial-coordinate}
satisfies
\[
 P^{-1}(2/w)^\ell\binom{m}{\ell}q|a|^\ell
 =O_t\bigl(P^{-1}(2/w)^\ell m^\ell(Pw/m)^\ell\bigr)
 =O_t(P^{\ell-1})=o_w(1)
\]
for every $\ell\ge 2$. Combining \cref{eq:ideal-radial}, the threshold
perturbation bound $C_t'\sum_{r=1}^t\rho_r$, and the concentration and
disjunction errors \cref{eq:uniform-linear,eq:OR-SUM}, we obtain for $2\le\ell\le t$:
\begin{equation}\label{eq:R-decomp}
 R_\ell(u)=\eta_t u_\ell + E_\ell(u),
 \qquad
 |E_\ell(u)|\le O_t(\delta)+C_t'\sum_{r=1}^t\rho_r+o_w(1).
\end{equation}

Next, for the individual Fourier residual $A_T(u)$ at $|T|=r\in\{1,\ldots,t\}$,
the expected contribution of the generic template lists and the $u_S$-independent
baseline of the threshold lists is invariant under coordinate permutations and
therefore cancels identically in $b_T-\bar b_r$. By \cref{eq:triangular}, the
level-$r$ threshold groups on $S=T$ contribute $\gamma_r(u_T-\bar u_r)$ to
$\E[b_T-\bar b_r]$, which combines with $+\bar u_r$ in
\cref{eq:angular-coordinate} to give the exact diagonal term $u_T$. For
$\ell<r$, the total influence on $A_T(u)$ of all controls $u_S$ at level
$|S|=\ell$ is bounded via \cref{eq:Kell,eq:triangular} by
\begin{align}\label{eq:leakage}
 2\binom{r}{\ell}\frac{\gamma_\ell}{\gamma_r}
                 \frac{(w)_{r-\ell}}{(m-\ell)_{r-\ell}}
 &\le(2+o(1))\frac{\rho_\ell}{\rho_r(r-\ell)!}
 \le(3+o(1))\frac{\rho_\ell}{\rho_r}.
\end{align}
Together with the concentration and disjunction-minus-sum errors
\cref{eq:uniform-linear,eq:OR-SUM} (which are $o_w(1)$ after dividing by
$\gamma_r$), this yields for every $T$ with $|T|=r\in\{1,\ldots,t\}$:
\begin{equation}\label{eq:A-decomp}
 A_T(u)=u_T+E_T(u),
 \qquad
 |E_T(u)|\le(3+o_w(1))\sum_{\ell<r}\frac{\rho_\ell}{\rho_r}+o_w(1).
\end{equation}

We now fix the constants $\delta>0$ and $\rho_t,\rho_{t-1},\ldots,\rho_1>0$
independently of $w$: choose $\delta>0$ so that the $O_t(\delta)$ term in
\cref{eq:R-decomp} is at most $\eta_t/8$, choose $\rho_t>0$ so that
$C_t' t\rho_t\le\eta_t/8$, and then choose $\rho_{t-1},\ldots,\rho_1>0$
inductively with $\rho_\ell\le\rho_t$ so that
\begin{equation}\label{eq:rho-choice}
 \sum_{\ell<r}\frac{3\rho_\ell}{\rho_r}<\frac14
 \qquad(2\le r\le t).
\end{equation}
Then for all sufficiently large $w$, \cref{eq:R-decomp,eq:A-decomp} hold with
\[
 |E_\ell(u)|<\frac{\eta_t}{2}\quad(2\le\ell\le t),
 \qquad
 |E_T(u)|<\frac12\quad(1\le|T|\le t),
\]
uniformly over $u\in[-1,1]^{d}$. This immediately yields the
strict boundary sign conditions on all opposite faces of
$[-1,1]^{d}$:
\begin{align}\label{eq:face-signs}
 R_\ell(u)&>0\quad(u_\ell=1),&
 R_\ell(u)&<0\quad(u_\ell=-1),\\
 A_T(u)&>0\quad(u_T=1),&
 A_T(u)&<0\quad(u_T=-1).\notag
\end{align}

Because each $L$-bit selector with $L=2w^2$ has $2^{2w^2}$ discrete cutpoints, we
partition $[-1,1]^{d}$ into a rectangular grid of mesh spacing at
most $2^{-2w^2}/(C_t N)$ along each coordinate axis. Stepping between adjacent
grid vertices changes at most $O_t(1)$ selector thresholds by $1$, altering the
Boolean function $G(u)$ on a set of inputs of $\U_m$-measure at most
\begin{equation}\label{eq:mesh-bound}
        \Delta_w\le C_t(N+m^t)2^{-2w^2}=2^{-2w^2+O_t(\log w)}.
\end{equation}
Along every grid edge, the normalized residuals $R_\ell$ and $A_T$
change by at most $2^{-2w^2+O_t(w)}$. Let $\widetilde R_\ell,\widetilde A_T$
denote the multilinear interpolation of the vertex values $(R_\ell,A_T)$ across
each grid cell. Because multilinear interpolation on each boundary face is a
convex combination of face vertices, $\widetilde R_\ell$ and $\widetilde A_T$
inherit the strict face signs \cref{eq:face-signs}. By the Poincar\'e--Miranda
theorem, there is a grid cell containing a common zero of
$(\widetilde R_\ell,\widetilde A_T)$. Any vertex $u^*$ of that grid cell defines
a monotone DNF $G_w=G(u^*)$ whose residuals satisfy
\begin{equation}\label{eq:tiny-residual}
       \max\left\{\max_{2\le\ell\le t}|R_\ell(u^*)|,
                  \max_{1\le|T|\le t}|A_T(u^*)|\right\}
       \le d\cdot 2^{-2w^2+O_t(w)}
       \le 2^{-2w^2+O_t(w)}.
\end{equation}

\paragraph{Deduction of product contact.}
Averaging $A_T(u^*)$ over $|T|=\ell$ gives $|\bar u_\ell^*|\le 2^{-2w^2+O_t(w)}$
because $\sum_{|T|=\ell}(b_T-\bar b_\ell)=0$. Hence
$|b_T-\bar b_\ell|\le\gamma_\ell 2^{-2w^2+O_t(w)}$ for all $|T|=\ell$, and
$|R_\ell(u^*)|\le 2^{-2w^2+O_t(w)}$ gives $|\bar b_\ell+qa^\ell|\le 2^{-2w^2+O_t(w)}$
for $2\le\ell\le t$ (while $\bar b_1+qa=0$ holds by definition of $a$). Since
$\E[(1-G_w(X))\chi_T(X)]=-b_T$ for $T\neq\varnothing$, dividing by
$q=\Prb(G_w(X)=0)$ yields, for every $T\subseteq[m]$ with $1\le|T|\le t$,
\begin{equation}\label{eq:product-Walsh}
   \left|\E[\chi_T(X)\mid G_w(X)=0]-a^{|T|}\right|
     =\left|\frac{-b_T}{q}-a^{|T|}\right|
     \le 2^{-2w^2+O_t(w)}.
\end{equation}
For any subset $S\subseteq[m]$ with $|S|\le t$ and any assignment $y\in\bits^S$,
Walsh--Fourier inversion gives
\begin{align*}
 \nu(X_S=y)&=2^{-|S|}\sum_{T\subseteq S}\chi_T(y)\,\E[\chi_T(X)\mid G_w(X)=0],\\
 \alpha_w^{|y|}(1-\alpha_w)^{|S|-|y|}
 &=2^{-|S|}\sum_{T\subseteq S}\chi_T(y)\,a^{|T|},
\end{align*}
where $\alpha_w=(1+a)/2$. Since $a=o(1)$, we have
$\alpha_w^{|y|}(1-\alpha_w)^{|S|-|y|}\ge 3^{-t}$, so the relative error in
\cref{eq:contact} is at most $3^t 2^{-2w^2+O_t(w)}\le 2^{-w^2}$ for all
sufficiently large $w$. Monotonicity of $G_w$ implies $\wh{G_w}(\{i\})\ge 0$ for
all $i$, so $\bar b_1\ge 0$, $a=-\bar b_1/q\le 0$, and $\alpha_w=(1+a)/2\le 1/2$.

\paragraph{Canonical term count.}
There are $O_t(N)$ active template copies. Each sparse pool has at most
$O_t(2^{\lceil cw\rceil})$ clauses, each binary selector has at most $L+1=2w^2+1$
clauses, and each threshold indicator $H_{S,r}^\pm$ has at most $2^t$ clauses.
Expanding their products produces a monotone DNF with at most
\[
 C_t N(L+1)2^{\lceil cw\rceil}=2^{cw+O_t(\log w)}\le p_w^{-C\log(t+2)}
\]
clauses (and no more canonical minterms after absorption), completing the proof
of \cref{thm:gate}.
\end{proof}

\section*{Acknowledgments}

\paragraph{Use of generative AI.}
ChatGPT (OpenAI) was used extensively in the mathematical development and
writing of this paper, including proposing proof arguments, simplifying
proofs, generating and revising manuscript text, and developing the
accompanying Lean formalization.


\begin{thebibliography}{99}

\bibitem{BFJ+94}
Avrim Blum, Merrick Furst, Jeffrey Jackson, Michael Kearns, Yishay Mansour,
and Steven Rudich.
\newblock Weakly learning DNF and characterizing statistical query learning
using Fourier analysis.
\newblock In \emph{Proceedings of the 26th Annual ACM Symposium on Theory of
Computing (STOC)}, pages 253--262, 1994.

\bibitem{BT96}
Nader H. Bshouty and Christino Tamon.
\newblock On the Fourier spectrum of monotone functions.
\newblock \emph{Journal of the ACM}, 43(4):747--770, 1996.

\bibitem{DLM+14}
Dana Dachman-Soled, Homin K. Lee, Tal Malkin, Rocco A. Servedio, Andrew Wan,
and Hoeteck Wee.
\newblock Optimal cryptographic hardness of learning monotone functions.
\newblock \emph{Theory of Computing}, 10(13):327--350, 2014.

\bibitem{Fel08}
Vitaly Feldman.
\newblock Evolvability from learning algorithms.
\newblock In \emph{Proceedings of the 40th Annual ACM Symposium on Theory of
Computing (STOC)}, pages 619--628, 2008.

\bibitem{Fel09}
Vitaly Feldman.
\newblock Robustness of evolvability.
\newblock In \emph{Proceedings of the 22nd Annual Conference on Learning Theory
(COLT)}, pages 277--292, 2009.

\bibitem{Fel12}
Vitaly Feldman.
\newblock Learning DNF expressions from Fourier spectrum.
\newblock In \emph{Proceedings of the 25th Annual Conference on Learning Theory
(COLT)}, volume 23 of \emph{JMLR Workshop and Conference Proceedings}, pages
17.1--17.19, 2012.

\bibitem{Fel17}
Vitaly Feldman.
\newblock A general characterization of the statistical query complexity.
\newblock In \emph{Proceedings of the 30th Annual Conference on Learning Theory
(COLT)}, volume 65 of \emph{Proceedings of Machine Learning Research}, pages
785--830, 2017.

\bibitem{FLS11}
Vitaly Feldman, Homin K. Lee, and Rocco A. Servedio.
\newblock Lower bounds and hardness amplification for learning shallow
monotone formulas.
\newblock In \emph{Proceedings of the 24th Annual Conference on Learning Theory
(COLT)}, volume 19 of \emph{JMLR Workshop and Conference Proceedings}, pages
273--292, 2011.

\bibitem{Jac97}
Jeffrey C. Jackson.
\newblock An efficient membership-query algorithm for learning DNF with respect
to the uniform distribution.
\newblock \emph{Journal of Computer and System Sciences}, 55(3):414--440, 1997.

\bibitem{KVV10}
Varun Kanade, Leslie G. Valiant, and Jennifer Wortman Vaughan.
\newblock Evolution with drifting targets.
\newblock In \emph{Proceedings of the 23rd Annual Conference on Learning Theory
(COLT)}, pages 155--167, 2010.

\bibitem{Kea98}
Michael Kearns.
\newblock Efficient noise-tolerant learning from statistical queries.
\newblock \emph{Journal of the ACM}, 45(6):983--1006, 1998.

\bibitem{KS04}
Adam R. Klivans and Rocco A. Servedio.
\newblock Learning DNF in time $2^{\tilde{O}(n^{1/3})}$.
\newblock \emph{Journal of Computer and System Sciences}, 68(2):303--318, 2004.

\bibitem{LMN93}
Nathan Linial, Yishay Mansour, and Noam Nisan.
\newblock Constant depth circuits, Fourier transform, and learnability.
\newblock \emph{Journal of the ACM}, 40(3):607--620, 1993.

\bibitem{Ser04}
Rocco A. Servedio.
\newblock On learning monotone DNF under product distributions.
\newblock \emph{Information and Computation}, 193(1):57--74, 2004.

\bibitem{Val84}
Leslie G. Valiant.
\newblock A theory of the learnable.
\newblock \emph{Communications of the ACM}, 27(11):1134--1142, 1984.

\bibitem{Val09}
Leslie G. Valiant.
\newblock Evolvability.
\newblock \emph{Journal of the ACM}, 56(1):3:1--3:21, 2009.

\bibitem{Ver90}
Karsten A. Verbeurgt.
\newblock Learning DNF under the uniform distribution in quasi-polynomial time.
\newblock In \emph{Proceedings of the 3rd Annual Workshop on Computational
Learning Theory (COLT)}, pages 314--326, 1990.

\end{thebibliography}
\end{document}